\documentclass[lettersize,journal]{IEEEtran}
\usepackage{amsmath,amsfonts}
\usepackage{algorithmic}
\usepackage{array}
\usepackage[caption=false,font=normalsize,labelfont=sf,textfont=sf]{subfig}
\usepackage{textcomp}
\usepackage{stfloats}
\usepackage{url}
\usepackage{verbatim}
\usepackage{graphicx}
\usepackage{multirow}
\usepackage{caption}
\usepackage[usenames,dvipsnames]{color}
\def\BibTeX{{\rm B\kern-.05em{\sc i\kern-.025em b}\kern-.08em
    T\kern-.1667em\lower.7ex\hbox{E}\kern-.125emX}}
\usepackage{balance}
\begin{document}
\title{Common Corruptions for Evaluating and Enhancing Robustness in Air-to-Air Visual Object Detection}

\author{Anastasios Arsenos, Vasileios Karampinis, Evangelos Petrongonas, Christos Skliros, Dimitrios Kollias, Stefanos Kollias, Athanasios Voulodimos
\thanks{Anastasios Arsenos is with the School of Electrical \& Computer Engineering, National Technical University Athens, Polytechnioupoli, Zografou, 15780, Greece (anarsenos@ails.ece.ntua.gr)}
\thanks{Vasileios Karampinis is with the School of Electrical \& Computer Engineering, National Technical University Athens, Polytechnioupoli, Zografou, 15780, Greece (vkarampinis@ails.ece.ntua.gr)}
\thanks{Evangelos Petrongona is with the School of Electrical \& Computer Engineering, National Technical University Athens, Polytechnioupoli, Zografou, 15780, Greece (vpetrog@microlab.ntua.gr)}
\thanks{Christos Skliros is with the Hellenic Drones S.A., Grigoriou Lampraki 17, Piraeus, 18533, Greece (c.skliros@hellenicdrones.com)}
\thanks{Dimitrios Kollias is with the School of Electronic Engineering \& Computer Science, Queen Mary University of London, UK (d.kollias@qmul.ac.uk)}
\thanks{Stefanos
Kollias is with the School of Electrical \& Computer Engineering, National Technical University Athens, Polytechnioupoli, Zografou, 15780, Greece (stefanos@cs.ntua.gr)}
\thanks{Athanasios Voulodimos is with the School of Electrical \& Computer Engineering, National Technical University Athens, Polytechnioupoli, Zografou, 15780, Greece (thanosv@mail.ntua.gr)}}

\markboth{Journal of \LaTeX\ Class Files,~Vol.~18, No.~9, September~2020}%
{How to Use the IEEEtran \LaTeX \ Templates}

\maketitle

\begin{abstract}
The main barrier to achieving fully autonomous flights lies in autonomous aircraft navigation. Managing non-cooperative traffic presents the most important challenge in this problem. The most efficient strategy for handling non-cooperative traffic is based on monocular video processing through deep learning models. This study contributes to the vision-based deep learning aircraft detection and tracking literature by investigating the impact of data corruption arising from environmental and hardware conditions on the effectiveness of these methods. More specifically, we designed $7$ types of common corruptions for camera inputs taking into account real-world flight conditions. By applying these corruptions to the Airborne Object Tracking (AOT) dataset we constructed the first robustness benchmark dataset named AOT-C for air-to-air aerial object detection. The corruptions included in this dataset cover a wide range of challenging conditions such as adverse weather and sensor noise. The second main contribution of this letter is to present an extensive experimental evaluation involving $8$ diverse object detectors to explore the degradation in the performance under escalating levels of corruptions (domain shifts). Based on the evaluation results, the key observations that emerge are the following: 1) One-stage detectors of the YOLO family demonstrate better robustness, 2) Transformer-based and multi-stage detectors like Faster R-CNN are extremely vulnerable to corruptions, 3) Robustness against corruptions is related to the generalization ability of models. The third main contribution is to present that finetuning on our augmented synthetic data results in improvements in the generalisation ability of the object detector in real-world flight experiments. \end{abstract}


\section{Introduction}
  
The visual identification of aerial vehicles has gained heightened interest in recent years as a fundamental technology for various critical applications. One such application is the implementation of see-and-avoid \cite{seeandavoidieeeTransactionsonIntelligentVehicles, senseandavoidletters} protocols among aircraft. Particularly noteworthy is the escalating presence of Advanced Air Mobility aircraft in low-altitude airspace, such as
those used for parcel delivery. Addressing the challenge of
ensuring the timely detection of other aerial objects is imperative
for safe navigation and collision avoidance in this context. Furthermore, the malicious deployment of micro UAVs has emerged as a significant menace to public safety and individual privacy in contemporary times. The visual detection of such malevolent micro UAVs constitutes a pivotal technology in the creation of civilian UAV defence systems \cite{AirtoAirMAV}. Another critical application is the visual detection of UAVs, a crucial component for the realization of vision-based UAV swarming systems \cite{swarm2018review}. In these systems, each aircraft is required to utilize onboard cameras for assessing the relative motion of its counterparts, underscoring the essential role of visual detection in enabling these technologies.

The categorization of aircraft detection encompasses two distinct application scenarios. The first scenario involves ground-to-air detection, where cameras placed on the ground are utilized to identify aircraft. The second scenario, which is the primary focus of this paper, pertains to air-to-air detection, where an aircraft leverages its onboard cameras to identify other aircraft. Although various sensor types, such as vision \cite{detFly}, radar \cite{radarbased}, lidar \cite{lidarbased} and RF-based \cite{rfbased} sensors, could be employed for aircraft detection, visual sensors emerge as one of the few viable choices for the air-to-air scenario, owing to the limited aircraft payload capacity. This paper specifically concentrates on the widely adopted RGB monocular cameras.


While deep learning methods have demonstrated exceptional performance in various object detection tasks, their potential for UAV detection has not been thoroughly explored or assessed thus far. In recent years, there has been a notable emergence of benchmark datasets \cite{detFly, AOT} specifically designed for air-to-air object detection and tracking. These datasets play a pivotal role in advancing the field by providing standardized evaluation frameworks and diverse scenarios for assessing the performance of detection and tracking algorithms in aerial environments.

Yet, the scenarios encompassed by these publicly available benchmarks are typically constrained. Notably, these datasets often lack representations of adverse weather effects, despite the potential impact of fog or rain on images as elucidated in studies \cite{fogintro, rainrender}. Beyond external scenarios, the internal noise inherent in sensors can further contribute to increased deviation and variance in ranging measurements. Ranging measurements, as highlighted in studies \cite{robustcvpr2023benchmarking, robustieee2023common}, can lead to data corruption and a subsequent decline in detector performance due to internal noise in the sensors. Therefore, it is imperative to conduct a thorough evaluation of an object detector under these varied corruptions before deploying it in real-world
environments.

A promising approach involves artificially introducing real-world corruptions to clean datasets to assess model robustness. For instance, ImageNet-C \cite{imagenetc} was initially introduced in image classification and includes $15$ corruption types, encompassing noise, blur, weather conditions, and digital corruptions. This methodology has been extended to $3D$ object detection and point cloud recognition \cite{robustieee2023common,robustcvpr2023benchmarking}. However, many of the examined corruptions are hypothetical and may not accurately reflect the challenges encountered in autonomous driving scenarios of aerial vehicles. Building a comprehensive benchmark for evaluating the robustness of air-to-air aerial detection under diverse real-world flying conditions remains a challenging task.

This paper takes the initial step toward establishing a robust approach to air-to-air aircraft detection by proposing the first robustness dataset comprising aerial object images and conducting a comprehensive experimental evaluation of representative deep-learning algorithms.  

The robustness assessment is paramount for air-to-air aerial object detection, especially in dynamic and complex environments. Aerial systems face various challenges, including adverse weather conditions, dynamic changes in lighting, and diverse backgrounds, making them susceptible to different sources of interference. Robustness assessment ensures that detection models can perform reliably and accurately under these challenging conditions that most likely have not been encountered in the training dataset (domain shift). Domain shift \cite{domainshift} represents a well-recognized challenge for learning algorithms, leading to unpredictable declines in performance under conditions divergent from those encountered during training. 

In aerial object detection \cite{detFly, AOT}, the deep learning models will be further affected by corruption compared to other autonomous driving tasks because flying objects have more degrees of freedom in their movement and they are faster than ground vehicles. In addition, interclass differences in aerial objects are smaller and blurrier than those in ground vehicles causing the decision boundary to rely more on details and thus is more susceptible to corruption. Moreover, aerial systems must be robust to corruption since their predictions may cause collisions. Therefore the robustness assessment is crucial in aerial object detection applications.




The novelty and contribution of this work are detailed as follows.

\begin{itemize}
    \item First, this paper introduces AOT-C, an air-to-air object detection robustness benchmark dataset for aerial objects. To the best of our knowledge, this is the first robustness benchmark for aerial object detection. 
    \item Second, based on the benchmark, we conduct extensive empirical studies to evaluate the robustness of $8$ diverse object detection algorithms to reveal their vulnerabilities under common corruptions.
    \item We demonstrate that fine-tuning our synthetically constructed corruptions improves the performance in real-world flight experiments that include challenging conditions and significantly differ from the training dataset (domain shift).  
\end{itemize}

\section{Related work}
\noindent This section gives a review of the existing studies on visual detection of aerial objects and benchmarking robustness using common corruptions. We only consider the case of using monocular cameras
and data-focused approach to robustness. We give an overview of some of the related topics within the constraints of space.

\subsection{Aerial Object Detection}

Until now, there has been a scarcity of comprehensive datasets tailored for training deep learning algorithms in aerial object detection. As a consequence, Opromolla and Fasano \cite{OPROMOLLA} created their own dataset to train the deep learning-based object detector. This dataset had a very limited number of images and shallow image resolution of $150 \times 150$ pixels, which resulted in poor detection accuracy. Recently, Zheng et al. \cite{detFly} (DetFly dataset) introduced a larger and more diverse dataset that includes various backgrounds and consists of $13271$ images of a target micro UAV (DJI Mavic). Lee et al. \cite{AirtoAirMAV} combined the DetFly dataset and their experimental dataset to boost the detector's performance that was based on YOLOv4-tiny model. 

Recent works \cite{airtrack, arsenos4674579nefeli} leveraged sophisticated deep learning-based object detectors and trackers, and relied on the Airborne Object Tracking (AOT) Dataset \cite{AOT} for training and evaluating their models. This dataset was introduced in $2021$ as part of the Airborne Object Tracking Challenge hosted by Amazon Prime Air. This dataset comprises approximately $5,000$ flight sequences,  resulting in a cumulative $164$ hours of flight data with over $3.3$ million labelled image frames. To the best of our knowledge, AOT dataset is the largest and most comprehensive dataset for aerial object detection and tracking. 



\subsection{Robustness Benchmarks}

Several studies \cite{imagenetc, ccwang2023robustness, cc3dkar20223d, wintr_R171} have demonstrated that Deep Neural Networks (DNNs) exhibit susceptibility to common corruptions. Especially, \cite{wintr_R171} suggested that detecting objects irrespective of image distortions or weather conditions is essential for practical implementations of deep learning, such as autonomous driving. To methodically explore and enhance DNN robustness against corruptions, corruption benchmarks were initially introduced in image recognition \cite{cchendrycks2019benchmarking} and later extended to 3D object detection \cite{robustcvpr2023benchmarking}, semantic segmentation \cite{ccsegment}, pose estimation \cite{ccpose}, and person Re-ID \cite{ccreid}. In addition, simulated imagery is used for air-to-ground object detection in \cite{he2023robustnessR172, konen2021increasedR173}.

However, numerous investigated corruptions are hypothetical and may not accurately represent real-world scenarios in autonomous navigation of UAVs. Moreover, in the context of air-to-air aerial detection, the objects are typically of small size. Hence, the artificially generated corruptions need to be crafted in a manner that ensures the object remains visible across all severity levels. As a consequence, developing a comprehensive benchmark for evaluating the robustness of air-to-air aerial object detection across diverse real-world flying conditions remains a challenging task. To the best of our knowledge, in this work, we introduce the first robustness benchmark dataset for aerial object detection.



\begin{figure*}[h!]
    \centering
    \includegraphics[width=0.71\textwidth]{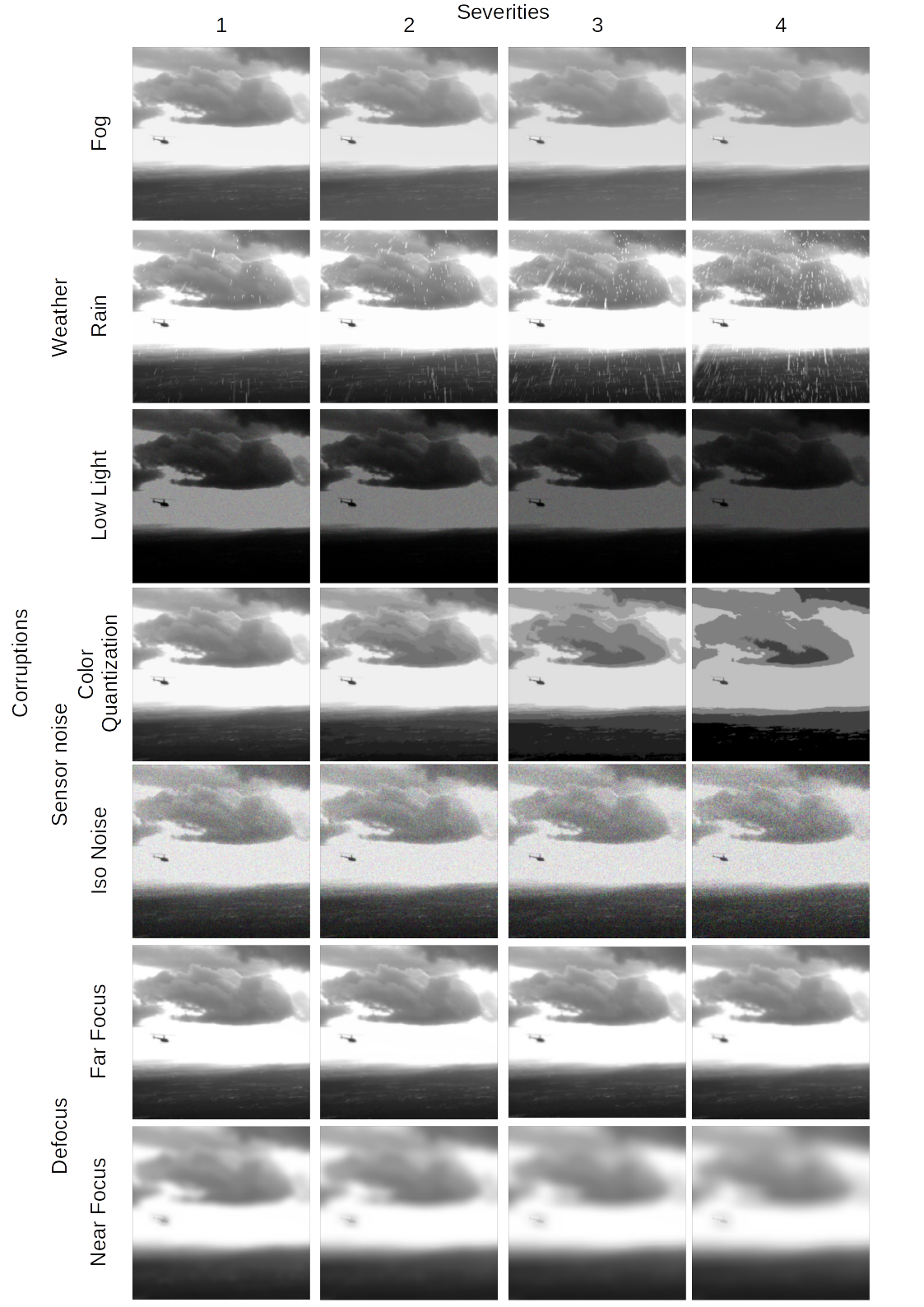}\hfill
    \caption{Visualization of the seven corruption types for each severity level in our benchmark}
    \label{fig:corsev}
    
\end{figure*}

\section{The Proposed Synthetic Dataset}

To conduct a comprehensive assessment of the corruption robustness of small object detection models (aerial object detection in our case), we establish a synthetic benchmark dataset using the widely used AOT dataset \cite{AOT}. The selected corruptions are then applied to the test set of this dataset, resulting in the AOT-C benchmark. It is worth noting that while some corruptions may naturally occur in a few samples of the datasets, we apply synthetic corruptions uniformly across all data. This ensures a fair comparison of model robustness under different corruptions and streamlines the process of data filtering.

\subsection{Corruption Setup}

Following the literature review, we illustrate seven corruption types in Fig \ref{fig:corsev} and classify them into three categories based on the typical presentations of common corruptions: weather, noise, and defocus. This dataset represents an initial endeavour, encompassing representative but not exhaustive corruptions. We encourage ongoing efforts to include a more diverse range of corruptions in future work. Brief introductions to each corruption pattern are provided below.

\textbf{Weather Corruptions}

Visual perception through cameras is susceptible to adverse weather conditions like rain and fog, where dense droplets of liquid or solid water can diminish the intensity of reflections and lower the signal-to-noise ratio (SNR) of received light. Moreover, floating droplets may produce false alarms and deceive sensors. These effects can significantly impact detectors in certain scenarios. To replicate three weather corruptions—rain, low-light (cloudy), and fog—we utilize simulators like \cite{rainrender} for rain and \cite{cc3dkar20223d} for fog and low-light conditions. For simulating low-light scenarios, we decrease pixel intensities and use Poisson-Gaussian distributed noise, mimicking imaging conditions in low-light settings based on \cite{cc3dkar20223d, poission_noise}.

\textbf{Sensor Noise Corruptions}

Noise corruptions arise due to constraints inherent in camera sensors. ISO noise follows a Poisson-Gaussian distribution, characterized by consistent photon noise (represented by a Poisson distribution) and varying electronic noise (represented by a Gaussian distribution) \cite{ccwang2023robustness, cc3dkar20223d}. Furthermore, we incorporate color quantization as an additional corruption, decreasing the bit depth of the RGB image \cite{cc3dkar20223d}.

\textbf{Defocus Corruptions}

Blurring due to defocus in a moving camera video may occur when the camera lens fails to achieve a sharp focus on objects within the scene. Several factors can contribute to this, such as abrupt changes in the distance between the camera and objects, swift movements of the camera, or constraints in the camera's autofocus system.

In scenarios involving a moving camera, particularly at high speeds or in situations with frequent depth of field changes, achieving precise focus on objects becomes challenging. Defocus blur is commonly observed in such cases, resulting in a lack of sharpness and clarity in parts of the video where objects may appear blurry or out of focus.

\subsection{Discussion on the disparity between synthetic and real-world corruptions}

Corruptions in the real world can stem from a multitude of diverse sources. For example, an autonomous UAV might experience adverse weather conditions and encounter uncommon objects simultaneously, leading to more intricate corruptions. While it is impractical to list all potential real-world corruptions, we systematically categorize seven corruptions into four levels, creating a practical testbed for controlled robustness evaluation.

In particular, for corruptions related to weather, we utilize state-of-the-art simulation methods that have been demonstrated to closely approximate real data \cite{rainrender, cc3dkar20223d, ccwang2023robustness}. Although an inevitable gap exists, we validate that the model's performance on synthetic weather aligns consistently with its performance on real data under adverse weather conditions by conducting real-world flight tests.

Additionally, each corruption type exhibits four severity levels, representing different intensities of manifestation. An example depicting four severity levels for each type of our synthesized corruptions is illustrated in Fig. \ref{fig:corsev}. These corruptions are implemented using functions, enabling seamless integration into the data loader for enhanced portability and storage efficiency.

 
\section{Experimental Setup}

\subsection{Benchmark Dataset}

For benchmarking air-to-air aerial object detection, we select 8 representative and diverse detectors: YOLOv5 \cite{yolov5}, YOLOv8 \cite{yolov8}, YOLOX \cite{yolox2021}, RetinaNet \cite{retinanet}, Faster R-CNN \cite{NIPS2015_14bfa6bb}, DiffusionDet \cite{chen2023diffusiondet}, DETR \cite{detr} and CenterNet2 \cite{CenterNet2} to cover different kinds of feature representations and proposal architectures. 
For a fair comparison, each detector in Table \ref{tab1} is trained with the clean training set of AOT, following the training strategy in each paper, and evaluated with the clean test set of AOT (first column of Table \ref{tab1}) and the corrupted test set of AOT (AOT-C) demonstrated in the second column of Table \ref{tab1}. 

According to the types of detection algorithms, the selected methods can be divided into two categories: one-stage networks (YOLOv5, YOLOv8, YOLOX, RetinaNet) and multi-stage networks Faster R-CNN,  DiffusionDet,  DETR, CenterNet2. One-stage object detectors, such as YOLO and RetinaNet, employ a single unified network to simultaneously predict object bounding boxes and classify their content in a single forward pass. In contrast,  Faster R-CNN uses a region proposal network (RPN) to generate potential bounding box proposals based on anchor boxes and then, these proposals are refined and classified by a subsequent network. DiffusionDet uses diffusion for detection. More particularly, in the training phase, object boxes transition from ground-truth boxes to a random distribution, and the model undergoes training to reverse this noise introduction process. During inference, the model iteratively refines a set of randomly generated boxes to produce the final output results. DETR, a deep learning model for object detection, utilizes the Transformer architecture, while CenterNet offers a probabilistic interpretation of the two-stage detection approach. This interpretation suggests the use of a strong first stage that learns to estimate object likelihoods.

All detectors are executed based on the open-source codes released on GitHub. For YOLOv5, YOLOv8 and YOLOX detectors we chose the large type model because is more suitable for small object detection \cite{yolomodelseval}.  
The training and evaluation are all executed on the NVIDIA RTX 4090 GPU with a memory of 24GB. The batch size of each detector is optimized to reach the limit of GPU memory. The default optimizers are employed, and parameters such as learning rate (LR), momentum, weight decay and batch size are meticulously adjusted through extensive experimentation.

\subsection{Real-world Flight tests}

Conducting real-world flight tests is crucial to showcase the model's performance under different conditions like camera resolution, surroundings and camera movement. Our real-world flight test consists of a domain shift compared to the training dataset (AOT) and also include adverse weather and lighting conditions. 

To provide further context, the real-flight dataset employed in our study encompasses $10,641$ images captured from two distinct drones equipped with different cameras. The first drone, a DJI Mavic, utilized its onboard camera to capture images at a resolution of $4K (3840\times2160)$ pixels, operating at a frame rate of $30$ frames per second (fps). The second drone, a custom model with approximate dimensions of $1000\times1000\times500$ mm ($width \times depth \times height$), utilized a GoPro Hero $8$ camera, capturing images at a resolution of $4000\times3000$ pixels, with a frame rate of $25$ fps. Both camera streams were processed at a uniform frame rate of $5$ fps to ensure consistency in data processing. 

While synthetic weather simulations provide a controlled environment for evaluating robustness, real-world flight tests introduce the complexities and nuances of actual flying conditions. These tests validate whether the model's performance on synthetic weather consistently aligns with its behaviour in real-world scenarios, including factors such as wind, precipitation, and atmospheric changes. By subjecting the model to authentic flight conditions, we ensure a more accurate and comprehensive evaluation, providing insights into the model's adaptability and reliability in dynamic, unpredictable environments that go beyond the constraints of synthetic simulations.

More particularly, these scenarios included variations in weather conditions such as adverse weather (e.g., cloudy-light rain), different lighting conditions, and changes in atmospheric factors that could impact object visibility. Additionally, the intruder behaviours were designed to be dynamic and unpredictable, involving maneuvers such as sudden changes in speed, direction, and altitude. The goal was to expose the object detector to challenging and unfamiliar situations, ensuring a robust assessment of its generalization ability in real-world, dynamic aerial environments.

In addition, we aim to illustrate the usefulness of our synthetically constructed corruptions in enhancing robustness. More particularly, we use the synthetic data corruptions as augmentations to fine-tune the object detection models to enhance their generalization ability on real-world flight data. Synthetic data corruptions serve as a means to simulate various challenging scenarios and environmental conditions that may be encountered during actual flights. The rationale behind this approach is that in challenging conditions (weather, noise defocus), the network adapts by prioritizing the most salient features relevant to object detection, thereby enhancing its robustness.
In addition, this strategy helps the model learn to recognize and handle distortions present in real-world flight data, ultimately leading to better generalization and performance when deployed in dynamic and unanticipated aerial environments.

\subsection{Metric setup}

The standard evaluation is performed on Aircraft, Helicopter and small UAV categories. These categories were aggregated into one category called Drone.  The evaluation metric is the Average Precision ($AP$) at an $IoU$ threshold of $0.5$. We denote model performance on the original validation (AOT's validation set) set as $AP_{clean}$. For each corruption type $c$ at each severity $s$, we adopt the same metric to measure model performance as $AP_{c,s}$,s. Then, the corruption robustness of a model is calculated by averaging over all corruption types and severities as

\begin{equation}
    AP_{cor} = \frac{1}{C} \sum_{c \in C} \frac{1}{4} \sum_{s=1}^{4} AP_{c,s}  
\end{equation}

where C is the set of corruptions in evaluation.

\begin{table}
\begin{center}
\caption{The benchmarking results of $8$ object detectors on AOT and AOT-C in terms of  Average Precision (AP), inference speed (fps) and model size (M)} 
\label{tab1}
\begin{tabular}{| c | c | c | c| c| }
\hline
Object detector & $AP_{clean} \uparrow$ & $AP\_{cor} \uparrow$ & fps $\downarrow$ & Model Size  $\downarrow$ \\
\hline
YOLOv5 \cite{yolov5}& 64.6 & \textbf{53.5} & 99 &46.5\\
\hline
YOLOv8 \cite{yolov8}& 56.4 & 41.2 & \textbf{110} & 43.7\\
\hline
YOLOX \cite{yolox2021}& \textbf{69.3}& 43.8&68 &54.2\\ 
\hline
RetinNet \cite{retinanet}& 35.7&20.0 &17& \textbf{37.9}\\
\hline 
FasterR-CNN \cite{fasterrcnn} & 52.9 &29.7 &15&41.3\\
\hline 
DiffusionDet \cite{chen2023diffusiondet}& 63.8& 35.7&30&110.5\\
\hline 
DETR \cite{detr}& 58.7& 26.1&27&41.2\\
\hline
CenterNet2 \cite{CenterNet2}& 66.2& 35.9&24&71.6\\
\hline
\end{tabular}
\end{center}
\end{table}

\begin{table*}[t]
\centering

\caption{The benchmarking results of $8$ object detectors on AOT-C. We show the performance under each corruption and the overall corruption robustness mAPcor averaged over all corruption types}
\label{tab:corruptions}
\begin{tabular}{|ll|l|l|l|l|l|l|l|l|}
\hline
\multicolumn{2}{|l|}{Corruption}                                  & YOLOv5 & YOLOv8 & YOLOX & RetinNet & FasterR-CNN & DiffusionDet & DETR & CenterNet2\\ \hline
\multicolumn{2}{|l|}{None ($AP_{clean}$)}                             &   64.6     & 56.4       &  69.3     &  35.7        &  52.9          &  63.8            &  58.7    & 66.2          \\ \hline
\multicolumn{1}{|l|}{\multirow{3}{*}{Weather}}     & fog          & 66.0  & 56.2       & 65.5 &  32.0   &  49.4     &    62.5     & 52.5      &54.0       \\ \cline{2-10} 
\multicolumn{1}{|l|}{}                             & rain         &  64.2      & 53.8       & 64.3      & 32.4         & 49.9           & 61.1             &   50.6   &   55.2        \\ \cline{2-10} 
\multicolumn{1}{|l|}{}                             & low light    & 49.4  &   33.4     & 38.4 &  18.8   &  21.0     &    24.0     & 22.5      &25.2       \\ \hline
\multicolumn{1}{|l|}{\multirow{2}{*}{Sensor noise}} & color\_quant & 49.8  &   41.2     & 42.3 &  19.8   &  35.0     &    38.9     &  10.8      &35.6       \\ \cline{2-10} 
\multicolumn{1}{|l|}{}                             & iso noise    & 32.5  & 18.3       & 20.2 &  4.9    &   8.2     &     9.3     &  6.3      &10.8        \\ \hline
\multicolumn{1}{|l|}{\multirow{2}{*}{Defocus}}     & far focus     & 58.3  & 50.2       & 51.6 &  25.0   &  38.4     &    48.3     & 38.7     &44.4       \\ \cline{2-10} 
\multicolumn{1}{|l|}{}                             & near focus   &   44.5     & 36.3       &  37.9     & 16.1          &   24.9         &  32.8            &  22.3    &  32.8         \\ \hline
\multicolumn{2}{|l|}{$AP_{cor}$}                           &   53.5     & 41.2       & 43.8      &  20.0        &  29.7          & 35.7             & 26.1     & 35.9          \\ \hline
\end{tabular}
\end{table*}


\section{Evaluation Results}
\noindent We present the corruption robustness of 8 object detection models on AOT-C in Table \ref{tab1}, where we present the results specifically for the Drone class, encompassing all airborne objects. The detailed outcomes for each corruption are available in Table \ref{tab:corruptions}. In general, corruption robustness exhibits a strong correlation with clean accuracy, with models like YOLOv5 and YOLOX achieving higher $AP_{cor}$ alongside higher $AP_{clean}$. This correlation is expected, given the consistent performance degradation observed across different models. The following analyses are based on the evaluation results.




\begin{figure*}[!t]
    \centering
    \subfloat[Base model evaluated on intense lighting\label{fig:first_case}]{\includegraphics[width=3.5in]{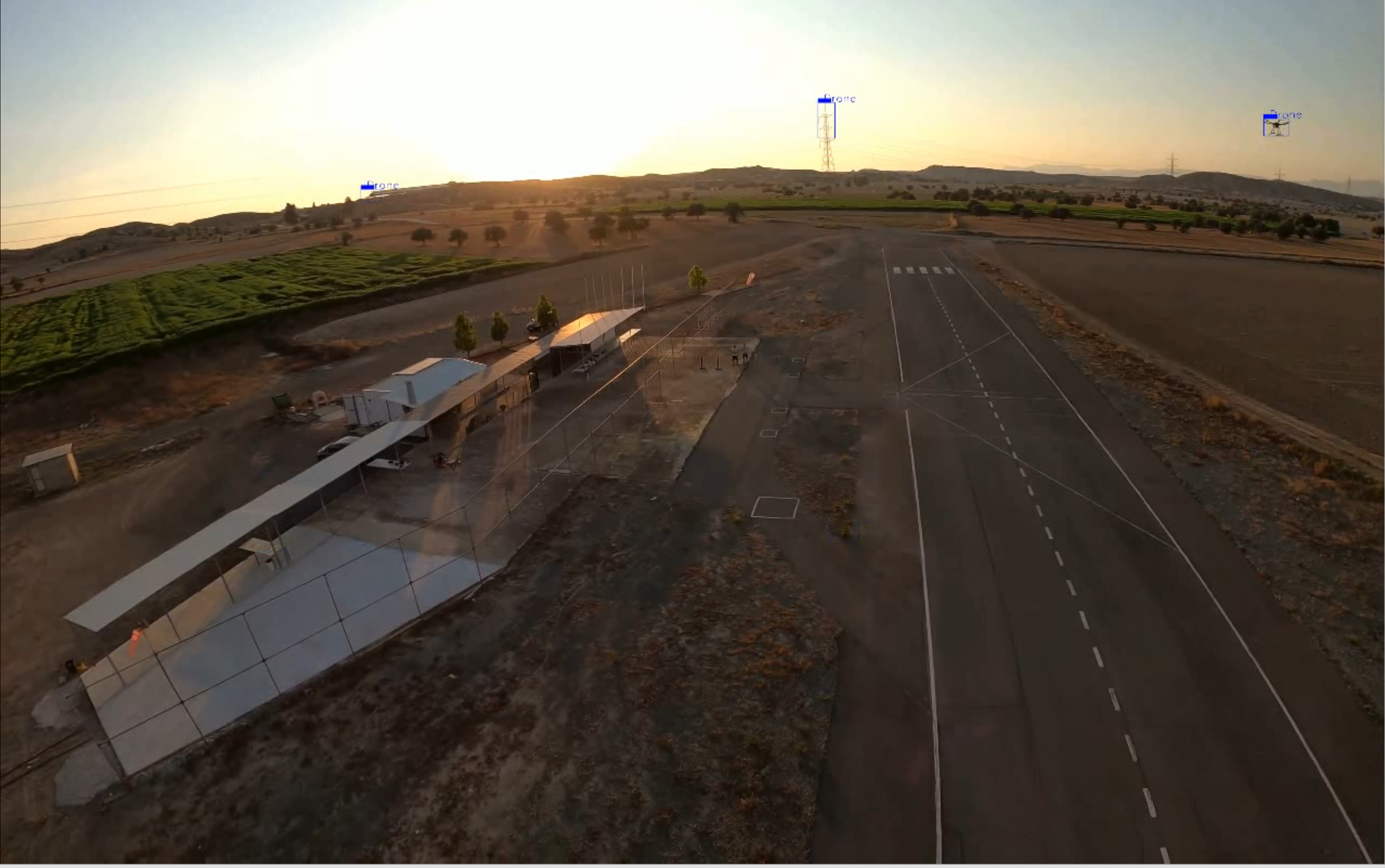}}
    \hfil
    \subfloat[Finetuned model evaluated on intense lighting\label{fig:second_case}]{\includegraphics[width=3.5in]{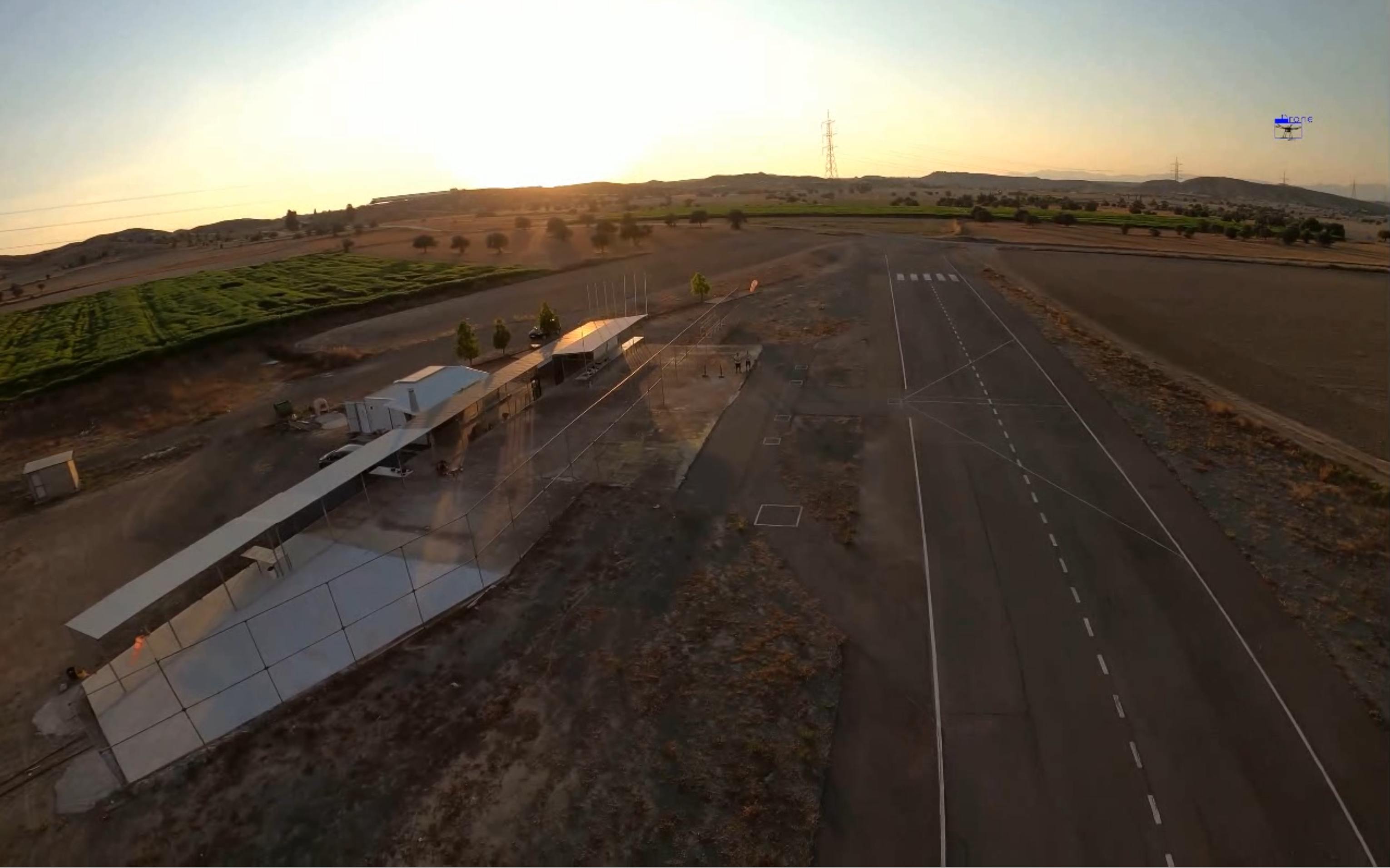}}
    \hfil
    \subfloat[Base model evaluated on cloudy-mild rain weather\label{fig:third_case}]{\includegraphics[width=3.5in]{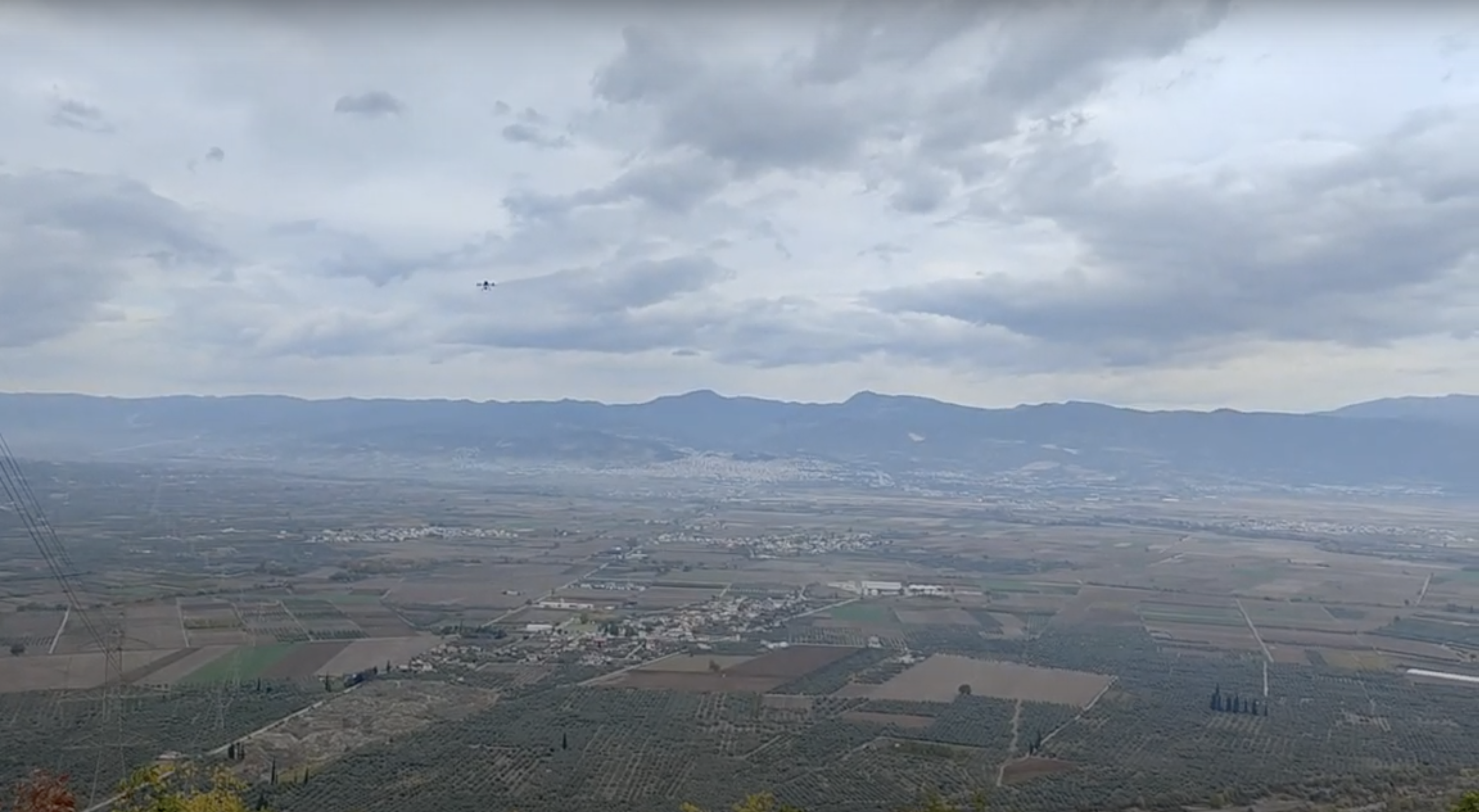}}
    \hfil
    \subfloat[Finetuned model evaluated in cloudy-mild rain weather\label{fig:fourth_case}]{\includegraphics[width=3.5in]{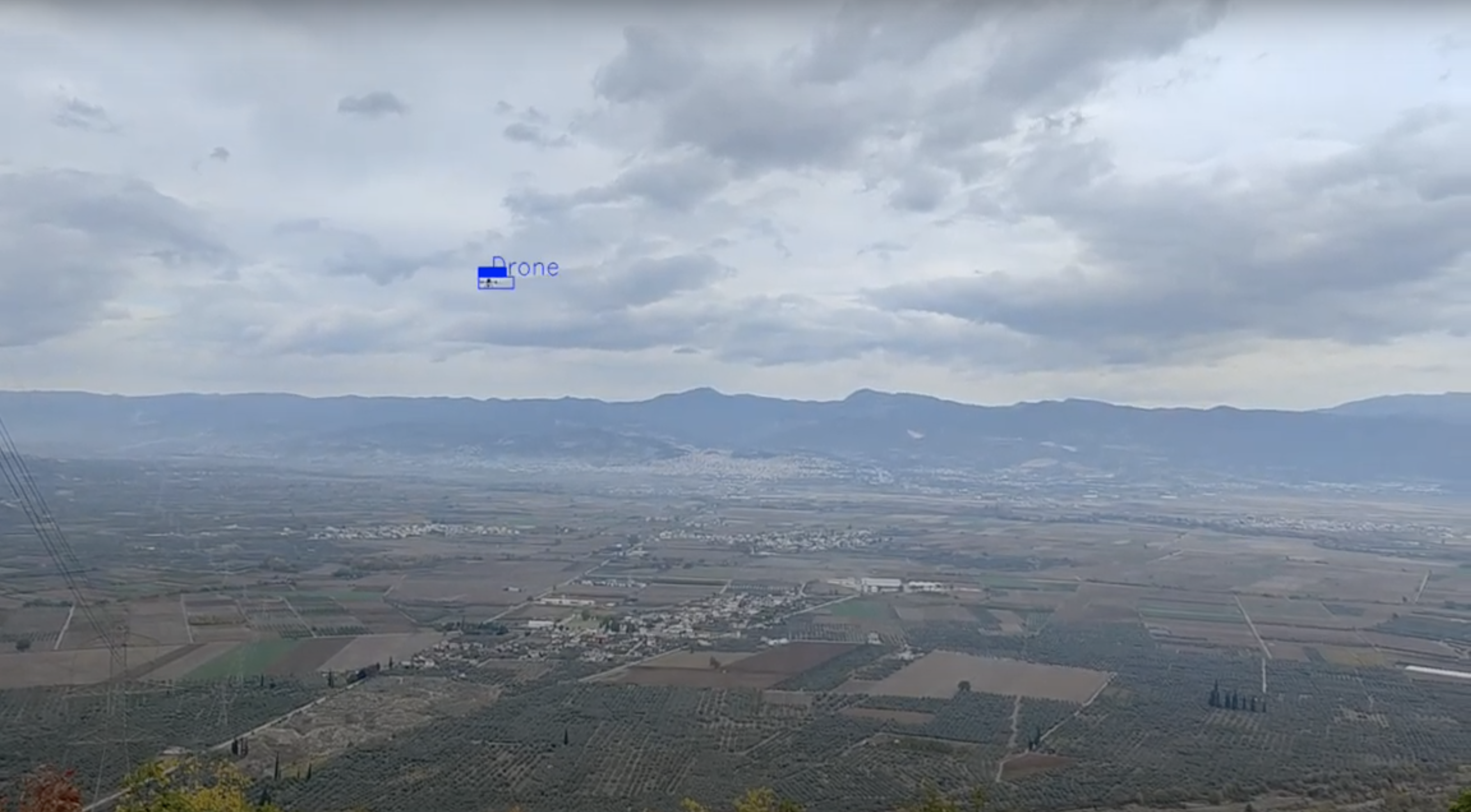}}
    \caption{Qualitative evaluation of YOLOv5 object detector on real flights when using corruptions as augmentation (Finetuned model) or not (Base model)} 
    \label{fig_sim}
\end{figure*}

\subsection{Robustness Evaluation on AOT Dataset}
Image corruptions decrease the prediction accuracy. The robustness performance of models measured in APs is shown in Table \ref{tab1}. YOLOv5 achieves the highest robustness against corruptions ($ AP_{cor}=51.4\%$) among all detectors, while RetinaNet is the worst ($20.0\%$). Among multi-stage networks, DiffusionDet is the most robust achieving $ AP_{cor}=35.7\%$), whereas DETR, which relies on a Transformer encoder-decoder architecture, achieves the worst ($26.1\%$). For one-stage networks, all YOLO family models perform well (more than $40\%$, whereas RetinaNet achieves only $20,0\%$. Although one-stage networks sacrifice detection performance to obtain high inference speed, the YOLO family models achieve better robustness on corrupted data and comparable performance in the original AOT evaluation set (clean) compared to multi-stage detectors like CenterNet2, DETR and diffusionDet. This suggests that YOLO models and especially YOLOv5 could be a good alternative for tasks requiring high computational efficiency and robustness to adverse conditions. 

In summary, YOLOv5 and YOLOX show stable and superior performance compared to the others. Since they could achieve higher computational speed and require fewer parameters (model size in millions of parameters), YOLOv5 and YOLOX may be good choice for tasks with limited computational resources.

To further evaluate the performance of the algorithms, we demonstrate the results for each of $7$ synthetically constucted corruptions separately in Table \ref{tab:corruptions}.

\subsection{Corruptions Types that Affect Aerial Detection}

How do different corruptions affect detectors’ overall accuracy? As shown in Table \ref{tab1}, the average $AP_{cor}$ is $22.7\%$ lower than the $AP_{clean}$ which is a significant decrease in the accuracy of detectors when confronted with various corruption patterns. These findings underscore the pressing need to tackle the robustness challenges faced by aerial object detectors.
Specifically, as indicated in Table \ref{tab:corruptions} iso noise, near focus and color quantization corruptions display the most AP loss among all other corruptions. This consists of a serious degradation in detection accuracy. By contrast, some corruption patterns (e.g. fog and rain) show fewer effects on detectors.
Our experimental study for each corruption separately in Table \ref{tab:corruptions} demonstrates that adverse weather does not significantly affect detectors’ accuracy. In contrast, sensor noise and especially iso noise considerably degrades the object detection performance of all models.



\subsection{Network Attributes Affecting Robustness}

As depicted in Table \ref{tab1}, multi-stage detectors exhibit lower robustness against common corruptions compared to one-stage detectors, as evidenced by their lower $AP_{cor}$. One potential explanation is that corrupted data may impact the proposal generation in the first stage (for both two-stage and one-stage detectors), and the poor-quality proposals can significantly affect the bounding box regression in the second stage (specifically for multi-stage detectors) \cite{robustieee2023common}. Furthermore, as indicated in Table \ref{tab:corruptions}, multi-stage detectors appear to be less susceptible to adverse weather conditions, displaying higher $AP_{cor}$, while being more vulnerable to sensor noise corruptions. Conversely, one-stage detectors, particularly those in the YOLO family, demonstrate more accurate results across all common corruptions.

\begin{table*}[t]
\caption{Evaluation results (measured in Average Precision (AP)) on AOT test set and on our real-world flight tests using corruptions as augmentations (Finetuned) and not (Base)}
\label{tab:ablation}
\centering
{\begin{tabular}{|c|c||c|c|c|c|c|c|c|c|}
\hline
Method  & Dataset  & {YOLOv5}  & {YOLOv8}  & {YOLOX}  & {RetinaNet}  & {FasterR-CNN} & {DiffusionDet} & {DETR}& {CenterNet2}\\
\hline
\hline

Base & AOT  & 64.6 & 56.4  & \textbf{69.3} & 35.7  & 52.9 & 63.8 & 58.7 & 66.2  \\
\hline
Finetuned &  AOT   &  65.6 &  56.1 & \textbf{66.4} & 36.1  & 51.2 & 62.4 & 58.1 & 64.3  \\
\hline
Base & real flights  & 37.3 & 24.8  & 32.7 & 14.2  & 29.0 & 28.5 & 33.4 & \textbf{38.9} \\
\hline
Finetuned &  real flights    & \textbf{48.6}  & 30.4  & 37.9 & 16.7 & 32.9 & 29.3 & 39.1 & 42.3  \\
\hline

\hline
\end{tabular}
}
\end{table*}

\subsection{Improving Generalisation Ability to real flights by finetuning on synthetic corruptions data}

We enhance the generalization ability of object detection models to real-world flight scenarios by employing a fine-tuning approach. This involves refining the models using synthetic corruptions as augmentation during the training process. The use of synthetic corruptions introduces diverse and challenging conditions to the training data, mimicking real-world variations such as adverse weather, noise, and defocus. By exposing the models to these augmented datasets, they learn to adapt and become more robust to a wider range of environmental factors, ultimately improving their ability to generalize and perform well in diverse and unanticipated situations encountered during real-world flight tests.

In Table \ref{tab:ablation} we show the results of the object detectors evaluated on the clean test set of AOT when we use synthetically corrupted data from the train set as augmentations (second row) and when we use the original train set (first row). It is observed that the finetuned models have almost the same results when evaluated on the same domain (AOT test set). Thus, this kind of augmentation does not affect the performance of the object detectors regarding the same dataset domain.

We also evaluated the finetuned and base object detectors on real-world flight tests that include variations in weather conditions such as cloudy, mild rain and different lighting conditions. Table \ref{tab:ablation} depicts that the models that were trained with synthetic corruptions (fourth row) achieve better robustness than the base models (third row). More particularly, finetuned YOLOv5 increases its performance by $11.3\%$ in terms of AP compared to the base YOLOv5. We also show in Fig \ref{fig_sim} some qualitative results of YOLOv5 model on our real-world flight tests. Images a and c have false positives and missed detection respectively while the finetuned model in images b and d showcase robust performance in these challenging scenarios.

\subsection{Robustness against corruptions is related to the generalization ability of models}

The robustness against corruptions is closely related to the generalization ability of object detection models. Generalization refers to a model's ability to perform well on new, unseen data, and this extends to handling variations and challenges present in real-world scenarios. Robustness against corruptions is a measure of how well a model maintains its performance in the face of unexpected variations, such as adverse weather conditions, noise, or other distortions.

This finding is illustrated by contrasting the $AP_{cor}$ values in Table \ref{tab1} with those of the base model assessed in real-world flight tests (third row) in Table \ref{tab:ablation}. Object detectors such as YOLOv5 and YOLOX, which demonstrated strong performance on AOT-C, also exhibit robust generalization to adverse real-world flight data. A notable exception to this trend is CenterNet, which, despite not performing exceptionally well on corrupted data, has outperformed all other detectors in real-world flight tests.


\section{Conclusion}

Computer vision models implemented in real-world scenarios will inevitably face distribution shifts that naturally occur compared to their training data. These shifts encompass various levels of distortions, including motion blur and changes in illumination, as well as semantic shifts such as object occlusion. Each of these represents a potential failure mode for a model and has been demonstrated to lead to significantly unreliable predictions [15, 23, 27, 31, 66]. Therefore, it is imperative to systematically test vulnerabilities to these shifts before deploying these models in real-world applications.

In this paper, we have conducted an extensive evaluation of cutting-edge object detection models concerning common corruptions. Through this investigation, we have uncovered various insights into the robustness of specific architectural configurations and the generalization capabilities of these models. These findings serve as valuable guidance for practitioners in selecting the appropriate model for their specific tasks, particularly when confronted with known types of image corruptions. We also demonstrated that our synthetically constructed corruptions can be used as augmentations to enhance the robustness of the object detectors when they are facing domain shifts. Additionally, our comprehensive analysis lays the groundwork for advancing the current state-of-the-art in robust air-to-air aerial object detection.

\bibliographystyle{IEEEtran}
\bibliography{IEEEfull}

\end{document}